\documentclass[11pt]{article}

\usepackage[final]{acl}

\usepackage{times}
\usepackage{latexsym}

\usepackage[T1]{fontenc}

\usepackage[utf8]{inputenc}

\usepackage{microtype}

\usepackage{inconsolata}

\usepackage{graphicx}

\usepackage{amsmath}

\usepackage{hyperref}
\usepackage{url}
\usepackage{booktabs}
\usepackage{multirow}
\usepackage{graphicx}
\usepackage{cleveref}
\usepackage[export]{adjustbox}
\usepackage{subfig}
\usepackage{pgfplots}
\pgfplotsset{compat=1.18}
\usepackage{comment}
\usepackage{subcaption}
\usepackage{csquotes}
\usepackage{euflag}

\title{Language Models Learn Universal Representations of Numbers\\and Here's Why You Should Care}

\author{Michal Štefánik$^{*\,1,2,3}$ \ \ Timothee Mickus$^{*\,2}$ \ \  Marek Kadlčík$^{*\,3}$ \\ \ \  \textbf{Bertram Højer}$^{4}$\vspace{2pt} \ \ \textbf{Michal Spiegel}$^{3,5}$ \ \ \textbf{Raúl Vázquez}$^{2}$ \ \ \textbf{Aman Sinha}$^{6}$ \\ \textbf{Josef Kuchař}$^{3}$ \ \ \textbf{Philipp Mondorf}$^{7,8}$\vspace{2pt} \ \ 
\textbf{Pontus Stenetorp}$^{1,9}$ \vspace{10pt}\\
$^{1}$R\&D Centre for Large Language Models, National Institute of Informatics, Japan\\
$^{2}$University of Helsinki \ \
$^{3}$TransformersClub @ Faculty of Informatics, Masaryk University\\
$^{4}$IT University of Copenhagen \ \
$^{5}$Kempelen Instutite of Information Technology\\
$^{6}$IECL-ATILF, Université de Lorraine - ICANS Strasbourg\\
$^{7}$MaiNLP, Center for Information and Language Processing, LMU Munich\\
$^{8}$Munich Center for Machine Learning (MCML) \ \ $^{9}$University College London\vspace{2pt}\\
$^{*}$\small{Equal contributors}
}

\begin{document}
\maketitle

\begin{abstract}

Prior work has shown that large language models (LLMs) often converge to accurate input embedding for numbers, based on sinusoidal representations.
In this work, we quantify that these representations are in fact strikingly systematic, to the point of being almost perfectly universal: different LLM families develop equivalent sinusoidal structures, and number representations are broadly interchangeable in a large swathe of experimental setups. 
We show that properly factoring in this characteristic is crucial when it comes to assessing how accurately LLMs encode numeric and other ordinal information, and that mechanistically enhancing this sinusoidality can also lead to reductions of LLMs' arithmetic errors.
\end{abstract}

\section{Introduction}


Mathematical problems provide an ideal playground in which to assess the current ability of large language models (LLMs) to manipulate precise information. 
While attempts to foster and improve mathematical reasoning in LLMs have been met with genuine empirical successes, e.g., as an aid to solving mathematical problems 
\citep{li2025provingolympiadinequalitiessynergizing}, this field is still rife with challenges, 
particularly in ensuring that LLMs tackling mathematical problems remain robust and reliable.
This calls for a deeper understanding of how robust the internal mechanisms of LLMs really are in operating over mathematical information.


Towards this goal, prior work identified that number representations stand out in terms of their sinusoidal structure \citep{nanda2023progress,NEURIPS2024_2cc8dc30,levy-geva-2025-language,kadlčík2025pretrainedlanguagemodelslearn}, accuracy of input embeddings \citep{kadlčík2025pretrainedlanguagemodelslearn}, and systematicity of internal mechanisms in addition \citep{kantamneni2025languagemodelsusetrigonometry,NEURIPS2024_2cc8dc30}.
While these works have provided pivotal insights as to how LLMs represent and manipulate numbers, their scope is limited: \citeauthor{kadlčík2025pretrainedlanguagemodelslearn}'s findings only pertain to input embeddings; \citeauthor{NEURIPS2024_2cc8dc30} find similarity of representations across models but do not quantify their alignment; \citeauthor{levy-geva-2025-language} focus only on decimals as one particular aspect of representations. Crucially, \textit{no} prior work considers \emph{natural language} contexts, which strongly curtails the generalizability and practical usefulness of their findings, needed for the development of tools able to interpret -- and verify -- LLMs in practice.

To address these gaps and create a holistic picture of how language models handle numbers, this paper presents two main findings.
First, we quantify that sinusoidal structures are \textbf{almost perfectly universal} across LLMs: models of different sizes and families converge to sinusoidal representations that are universal across \emph{models}~(§\ref{sec:universal:inputs}) and \emph{layers}~(§\ref{sec:universal:x-layers}) while remaining highly systematic and \emph{accurate} throughout the model's internal processing~(§\ref{sec:universal:contexts}) -- showing it is possible to track the processing of numeric information throughout models.

Second, we demonstrate the necessity of \textbf{suitable explainability tools} to make sense of how LLMs process numbers: we show how accounting for the sinusoidal nature of representations allows us to unravel the mechanism of manipulating large, multi-token numbers~(§\ref{sec:multitok}) as well as other ordinal data~(§\ref{sec:ordinal}) in LLMs, and we mechanistically prove that a precision of sinusoidality in the model's internal representation determines its output accuracy across arithmetic tasks~(§\ref{sec:error-tracking}).

In summary, our main contributions are:
\begin{enumerate}\setlength{\itemsep}{0pt}
    \item We develop a new, more accurate probe of numeric values that generalizes well across different tasks and models
    \item We quantify that numeric representations of numbers are sinusoidal and highly accurate \textit{throughout the model}, beyond the input embedding representations; 
    \item We are the first to analyze the LLMs' numeric processing in \textit{natural-language} settings and show that training numeric probes with real-world data is crucial for creating generalizing, practically useful probes;
    \item We investigate mechanisms of numeric manipulations beyond addition and provide a mechanistic evidence that sinusoidality is crucial for models' accuracy. 
\end{enumerate}
\makeatletter\ifacl@finalcopy 
We make the new probe, with all our methods and analyses publicly available for any use.\footnote{\url{https://github.com/prompteus/numllama}}
\fi\makeatother



\section{Related work}
\label{sec:background}


\paragraph{Mathematical capabilities of LLMs} The seminal findings of \citet{brown2020languagemodelsfewshotlearners} on remarkable arithmetical abilities of LLMs have ushered in a large body of work aimed at evaluating more complex mathematical capabilities of LLMs \citep{hendrycks2021measuring,cobbe2021trainingverifierssolvemath,sun-etal-2024-benchmarking,yu2024metamath} or lack thereof  \citep{nikankin2025arithmeticalgorithmslanguagemodels}.
This in turn has encouraged a focus on how to bolster the capabilities of LLMs or interpret how LLMs perform computations \citep{zhang2024interpretingimprovinglargelanguage,stolfo-etal-2023-mechanistic}. Most relevant to ours, some of the previous work focuses specifically on how numbers should be represented in principle.
For instance, \citet{charton2022lingebratransformers} assesses an impact on models performance in linear algebra problems when employing different encoding schemes based on scientific notation on linear algebra problems.
\citet{golkar2023xval} propose to encode numeric values by incorporating a scaled, learned control token \texttt{<NUM>}.
\citet{feng2024numericalprecisionaffectsmathematical} remark that the precision of the numeric representation type impacts performance on arithmetic tasks, and argue to use the logarithmic-precision architecture of \citet{feng2023towards}.

\paragraph{Numeric representations in LLMs} The increased focus on evaluating LLM performance in verifiable domains such as math problems has also led to increased interest in the mechanisms by which LLMs compute arithmetic functions.
\citet{nanda2023progress} demonstrate how a model trained from scratch on modular addition relies on trigonometric operation: it maps inputs onto a unit circle, corresponding to a specific rotation, and then learns to combine the two rotations to derive a valid solution.
\citet{kantamneni2025languagemodelsusetrigonometry} apply circuit analysis to understand how general pretrained models perform addition, and report that representations are mapped onto a helix that can be manipulated using trigonometric operations.
\citet{NEURIPS2024_2cc8dc30} refine these observations in qualitative analysis of Fourier components in LLMs' representations (note that Fourier transformations map arbitrary functions into combinations of periodic trigonometric functions), and highlight the distinct role of attention and feedforward sublayers.
Focusing more narrowly on representations rather than processing, \citet{zhu-etal-2025-language} argue that non-linear probes do not provide a better fit than linear probes.
\citet{levy-geva-2025-language} remark that it is possible to retrieve the digit (value mod 10) of numeric inputs with high accuracy.
Our work builds upon \citet{kadlčík2025pretrainedlanguagemodelslearn}  that propose a probe architecture that factors in the sinusoidal nature of representations, and show that they can retrieve numeric values from input embeddings with much higher accuracy.


\section{Universality of Sinusoidal Representations of Numbers}
\label{sec:universal}

\subsection{Different Models Learn Equivalent Representations of Numbers}
\label{sec:universal:inputs}

The first point we address is the distinctiveness of LLMs' representations of numbers.
Results from \citet{kadlčík2025pretrainedlanguagemodelslearn} and \citet{zhu-etal-2025-language} suggest that different models converge to the same \emph{type} of number representations, but they do not provide a direct assessment of how \textit{closely} the representations match across models -- which could evidence that a shared representation is a causal consequence of architectural bias and optimization process rather than a coincidental artifact.
Following \citeauthor{kadlčík2025pretrainedlanguagemodelslearn}, we focus on the input embeddings from eight LLMs of diverse sizes and families with open-sourced, pre-trained checkpoints: OLMo 2 \citep{olmo20252olmo2furious}, Llama 3 \citep{grattafiori2024llama3herdmodels} and Phi 4 \citep{abdin2024phi4technicalreport}.

\begin{figure}[t!]
    \centering
    \vspace{-0.25cm}
    \subfloat[Cosine-based RSA for \\ number tokens]{
        \includegraphics[height=0.2125\textwidth, trim={1cm 1cm 3.5cm 0}, clip]{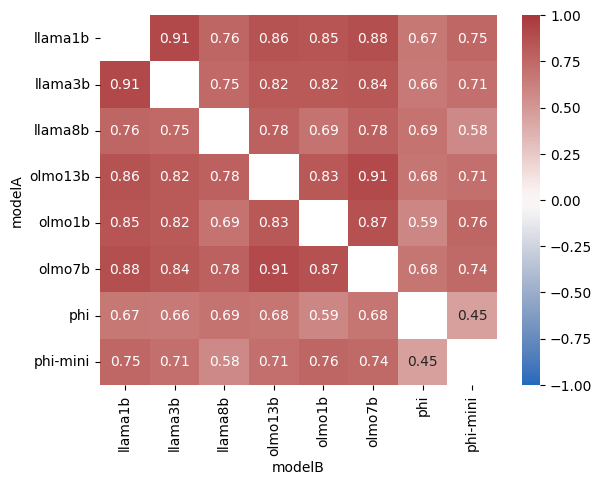}
    }~~
    \subfloat[Cosine-based RSA for \\ random tokens]{
        \includegraphics[height=0.2125\textwidth, trim={3.55cm 1cm 0 0}, clip]{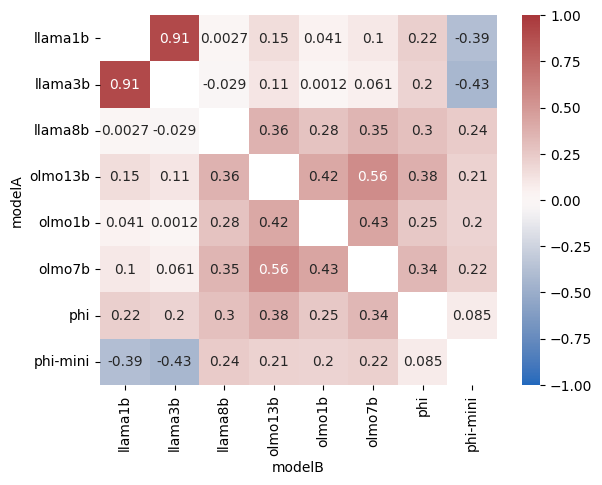}
    }
    \caption{RSA scores for input embeddings of LLMs}
    \label{fig:rsa similarity}
\end{figure}

To quantify whether models converge to similar embeddings, we start by conducting a simple Representational Similarity Analysis (RSA; \citealp{Kriegeskorte2008-vs}) across the input embeddings for number word-pieces in paired models. 
As a baseline, we compute RSA scores for a random sample of 1000 pieces present in the vocabularies of all the models under consideration.
Results of the analysis are displayed in \Cref{fig:rsa similarity}. Across all pairs of models, we find that number embeddings systematically yield consistently higher RSA scores than a random sample of word pieces. While some models (especially in the Phi family) are not as well aligned with other models, we do observe high scores both within and across families, demonstrating that models converge to embedding spaces with \textbf{equivalent similarity structure}.

\begin{figure*}[ht!]
    \centering
    \subfloat[\label{fig:fft iou:example}Contributions of Fourier base frequencies (shown example: Llama-3 1B)]{
        \raisebox{2.2ex}{\includegraphics[max width=0.3\textwidth]{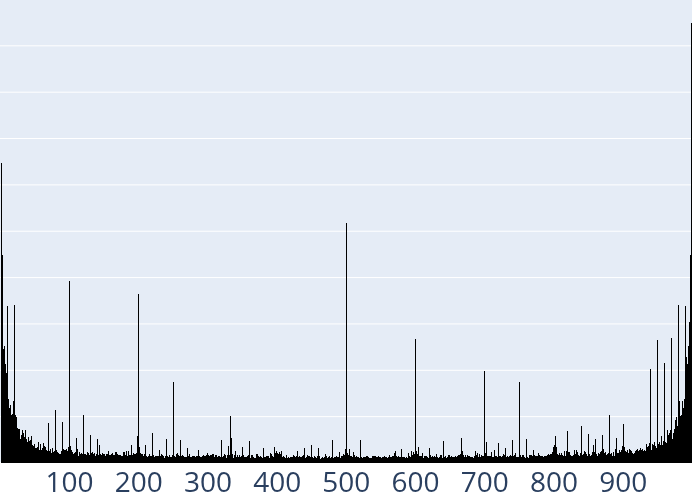}
    }}~
    \subfloat[\label{fig:fft iou:num}IoU for number pieces]{
        \includegraphics[max width=0.3\textwidth]{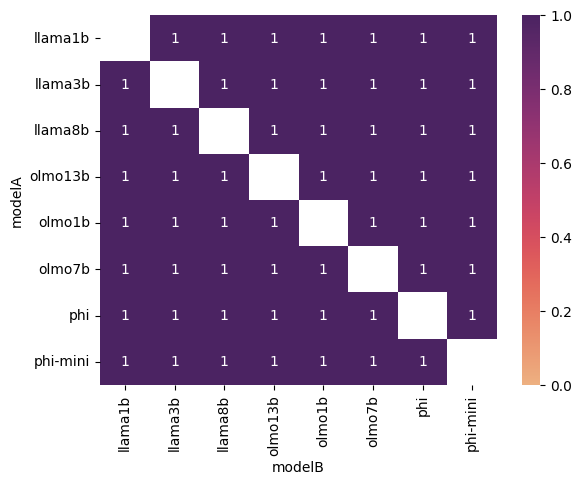}
    }~
    \subfloat[\label{fig:fft iou:rand}IoU for random pieces]{
        \includegraphics[max width=0.3\textwidth]{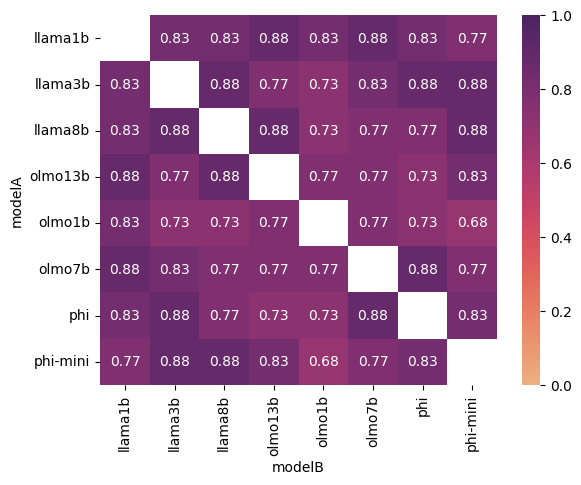}
    }
    \caption{Intersection-over-union of top $k=63$ Fourier base frequencies}
    \vspace{-10pt}
    \label{fig:fft iou}
\end{figure*}
Another approach to quantifying the similarity of number representations is to inspect their base frequencies through Fourier decompositions \citep{kadlčík2025pretrainedlanguagemodelslearn,NEURIPS2024_2cc8dc30}.
Applying a PCA transformation followed by a Fourier transform allows us to quantify the \textit{magnitude} of individual frequencies (\Cref{fig:fft iou:example}). 
Subsequently, we determine whether two models agree as to how they rank frequencies by computing a simple intersection-over-union (IoU) of the top $k$ frequencies.
We repeat this process for every pair of models, using the input embeddings of number pieces (\Cref{fig:fft iou:num}) as well as that of random pieces (\Cref{fig:fft iou:rand}).
When considering the top $k=63$ frequencies, we find \textbf{perfect agreement across all models} for numbers. This provides a quantitative basis for some of \citeauthor{NEURIPS2024_2cc8dc30}'s (\citeyear{NEURIPS2024_2cc8dc30}; \S C.1) qualitative observations.\footnote{See our \Cref{appendix:fft threshold} for technical details, including how the value of $k$ is set.} 

In summary, our analyses show that the LLMs from different families learn to represent numbers in the same topology and trigonometric features using the same dominant frequencies.
This is specific to numbers: we do not reproduce these observations with other word pieces shared across models.

\subsection{Number Representations are Always Sinusoidal}
\label{sec:universal:contexts}

In \Cref{sec:universal:inputs}, we presented evidence that the input embeddings of numbers \textit{across} different LLMs are highly consistent. We now focus on whether the sinusoidal mechanism of representations is maintained and employed \textit{throughout} the model.

For qualitative assessment, we visualize the PCA and its Fourier transform of internal activations of numerical tokens in \Cref{appendix:pca_fourier}. Besides the apparent wave-like pattern, the representations have a sparse Fourier transform, confirming the sinusoidal character.
For quantitative analysis, we look at the model's internal representations through the lens of the sinusoidal probe proposed by \citet{kadlčík2025pretrainedlanguagemodelslearn}. This probe was designed to map input embeddings of LLMs into an integer, thus classifying embeddings into predefined range of numeric values. It is defined as:
\begin{equation}
\resizebox{0.95\columnwidth}{!}{$\begin{aligned}%
    f_{\sin}(\mathbf{x}) &= (\mathbf{W}_\mathrm{out}\mathbf{S})^T(\mathbf{W}_\mathrm{in}\mathbf{x}) \\
    \mathbf{S}_{ij} &= \begin{cases}
        \sin(i e^j 1000 / h) \hfill \mathrm{if~} j \equiv 0 \mod 2 \\
        \cos(i e^{j+1} 1000 / h) ~ \mathrm{if~} j \equiv 1 \mod 2
    \end{cases}
\end{aligned}$}%
\label{eq:sin}
\end{equation}
\noindent where $\mathbf{W}_\mathrm{in}: h \times d$ and $\mathbf{W}_\mathrm{out}: h \times d$ are learned parameters, and $\mathbf{S} : h \times 1000$ injects an inductive bias in the classifier towards sinusoidality.\footnote{Unless otherwise stated, we use $d=100$; $h$ corresponds to the inner dimensionality of the LLM at hand.}

While \citeauthor{kadlčík2025pretrainedlanguagemodelslearn} demonstrate that such a probe properly benefits from the expected sinusoidal inductive bias and greatly bolster accuracy for number representations, their approach does not enforce a strict sinusoidal structure and may fail to generalize due to redundancy in a model's representation. As an alternative, we design a new probe which assigns a \emph{single sine wave to each latent feature} and \emph{directly} optimize the frequency, phase-shift, and amplitude of each sine wave:
\begin{equation}
\begin{aligned}
    f_{\mathrm{param}\text{--}\sin}(\mathbf{x}) &= \mathbf{S}^T(\mathbf{W}_\mathrm{in}\mathbf{x})\\
    \mathbf{S} &= \mathbf{w}_a \sin( \mathbf{w}_\phi + \mathbf{w}_f \mathbf{r}^T ) 
\end{aligned}
\label{eq:new-sin}
\end{equation}
where $\mathbf{r}$ are constant label values ($0...999$ for all models we analyze), and $\mathbf{w}_f, \mathbf{w}_a, \mathbf{w}_\phi \in \mathbb{R}^d$ are three vectors of learnable parameters modeling the frequencies, amplitudes and the phase-shifts of the sine waves.
This architecture \textit{strictly constrains} the classification matrix $\mathbf{S}$ to take the form of $d$ sine waves sampled at $n$ evenly-spaced points. The configuration of initialization and hyperparameters is provided in Appendix \ref{appendix:datasets}.


We first assess whether sinusoidal probes are the most suitable, i.e., an accurate choice for decoding the \textit{internal} representations of the model.  
To contextualize the accuracy in terms of sinusoidal quality of the representation,  we also evaluate other types of probes used to decode numbers in previous work \cite{feng2023towards,zhu-etal-2025-language,kadlčík2025pretrainedlanguagemodelslearn} as baselines. 
We fit separate probes for activations from every layer across six different language models from three families.
Contrary to previous work, we assess the representations of numbers in \textit{natural-language} contexts, rather than as single tokens or synthetic arithmetical commands.
Our contexts cover four domains (temporal, arithmetic, medical and culinary) covering numerical contexts including quantities, measurements or timestamps (detailed in Appendix~\ref{appx:natural_lang_contexts}).
To assess generalization rather than memorization capacity, we \textit{split} the prompts and corresponding representations into training, validation and test sets containing \textit{distinct} extracted numbers ($x_2$).

\begin{figure}[ht!]
    \centering
    \subfloat[\label{fig:per_layer_acc_across_probes} using probes from prior work with different bases]{%
        \includegraphics[width=0.9\columnwidth]{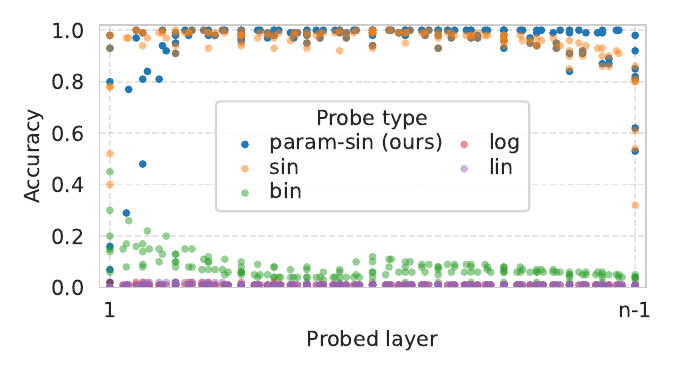}%
    }
    
    \subfloat[\label{fig:per_layer_acc_across_models_natural} across different language models; param-sin probe]{%
        \includegraphics[width=0.9\columnwidth]{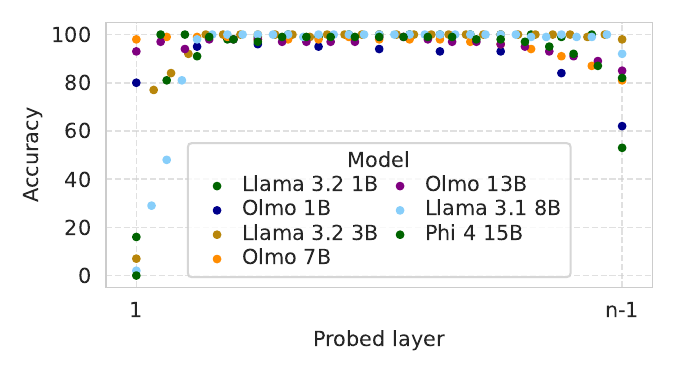}%
    }
    \caption{\textbf{Accuracy of decoding numbers} from internal activations of language models}
    \vspace{-10pt}
    \label{fig:perf-l2l}
\end{figure}

\Cref{fig:per_layer_acc_across_probes} shows the accuracy of different probes extracting the value of input numbers from representations of each model's layer, evaluated across six different models.
As we can see, both sinusoidal probes reach over 80\% of accuracy in all but two cases and over 90\% of accuracy for a majority of probing scenarios.
Here, the superior accuracy of both sinusoidal probes provides an indicator of both (i) the sinusoidal character, and (ii) the accuracy of input number representation across layers --- the best probe results present a lower bar of the accuracy of numeric representation on a given layer.

\Cref{fig:per_layer_acc_across_models_natural} disentangles the accuracies of the param-sin probe from \Cref{fig:per_layer_acc_across_probes} by the model, covering six LLMs of different sizes and three different families. 
While the probing is slightly less accurate on some particular layers of particular models, for none of the analyzed models does the sinusoidal probe systematically underperform.
If we take into account that our probes need not be perfect and reflect a \emph{lower bound} of real accuracy of numeric representations, these results taken together show that a sinusoidal representation is a decisively dominant scheme of numeric representations in language models, corroborating and extending the observations of \citet{kantamneni2025languagemodelsusetrigonometry}.

A vast majority of cases performing lower than 80\% occurs in probing the models' first and a last layer; in \Cref{appendix:sparsity}, we analyze these cases in detail through the lens of probes learned representations and find the substantial qualitative difference in the \textit{sparsity} of representations of numbers. 
While the middle layers of all models resort to a `dense' representation decomposed into a small set of features bearing full numeric information, the initial and final layers scatter the numeric information across most of the latent features. 
As the scattered features share the information with natural language, a robust recovery in a realistic setting of natural contexts becomes more difficult.

\paragraph{Generalization across contexts}
We can see that sinusoidal probes trained with natural-language contexts can provide highly accurate estimates of LLMs' numerical states across models.
However, favoring simplicity, all of the previous work (§\ref{sec:background}) investigated numerical representations in \textit{synthetic} mathematical contexts.
To assess whether such a simplification comes for a price in practical applicability of resulting probes, we explore whether the choice of training contexts affects generalization of resulting probes.


\begin{figure}[th!]
    \centering
    \includegraphics[max width=\columnwidth]{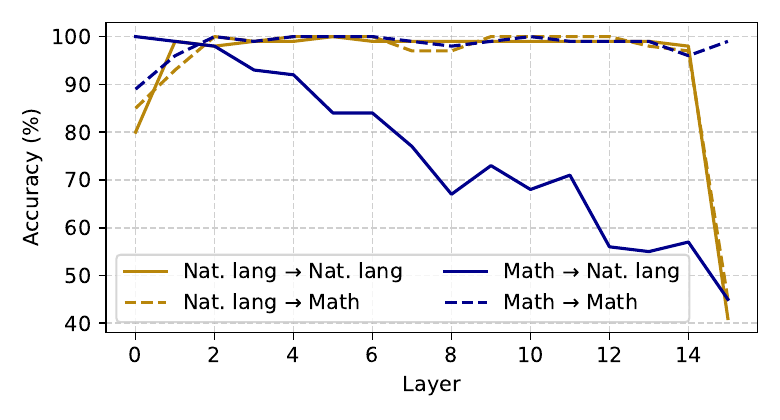}
    \vspace{-25pt}
    \caption{\textbf{Generalization of probes} fitted on representations in natural language (yellow), and synthetic, mathematical contexts (blue), evaluated on natural-language (solid) and math contexts (dashed). Llama 3.2 1B.}
    \label{fig:nat lang}
\end{figure}

In \Cref{fig:nat lang}, we compare the performance of probes trained on natural-language contexts (yellow) and synthetically-constructed math contexts (blue), evaluated on natural-language inputs (solid lines) and synthetic inputs (dashed lines). 
We can see that probes fitted with natural-language contexts are much more robust in application across \textit{both} math and natural-language context.
While both probes generalize well within their training context type, \textbf{only the natural-language probes generalize across evaluation settings}, robust to the context's distribution shift.
This observation has two important implications. First, it informs future work aiming to explain models' internal computations in practical applications to favor probes trained in natural-language settings. Second, it restrains future work in interpretability and explainability from drawing general conclusions on models' mechanics from fully-synthetic settings, evidencing that such conclusions may not generalize to real-world applications.


In summary, our results underscore that the same type of sinusoidal representation of numbers holds in general across different model types. 
Numbers are represented in a similar, systematic fashion, regardless of which layer or type of context we consider. 
While we can expect this to be in part due to the residual stream across layers, we note that there is no a priori reason for an LLM to preserve sinusoidality across layers, and previous work shows that input tokens tend to have a negligible impact on upper layer representations \citep{mickus-etal-2022-dissect}.

\subsection{Different Layers use Interchangeable Numeric Representations}
\label{sec:universal:x-layers}

Having found the remarkable degree to which the embeddings for numbers align across different models and the noteworthy universality of sinusoidal representations in varied contexts, as a final step, we turn to assess whether the models utilize the identical representation throughout its computation.
To investigate that, we employ the probes from  \Cref{sec:universal:contexts} trained separately for each separate model layer ($L_i$) and measure the accuracy of each probe on representations derived for every \textit{other} layer $L_j \neq L_i$.

While informative, this approach still comes with the caveat that probes trained for a given layer $L_i$ may pick up on idiosyncrasies inherent to a specific layer --- i.e., the probes might not disentangle what is specific to numbers as opposed to what is specific to $L_i$. To address this point, we also fit probes using \textit{all but one} layer ($L_1 , L_2, \dots L_{i-1}, ~ L_{i+1}, \dots L_n $), and evaluate performance on representations from the held out layer $L_i$.
As before, we hold out 100 numbers for assessing generalization in validation and test conditions.


\begin{figure*}[ht!]
    \centering
    \subfloat[\label{fig:one-v-one}accuracy of probes trained for Llama 1B on each single layer (rows) evaluated on other layers (columns)]{
    \includegraphics[width=0.275\textwidth]{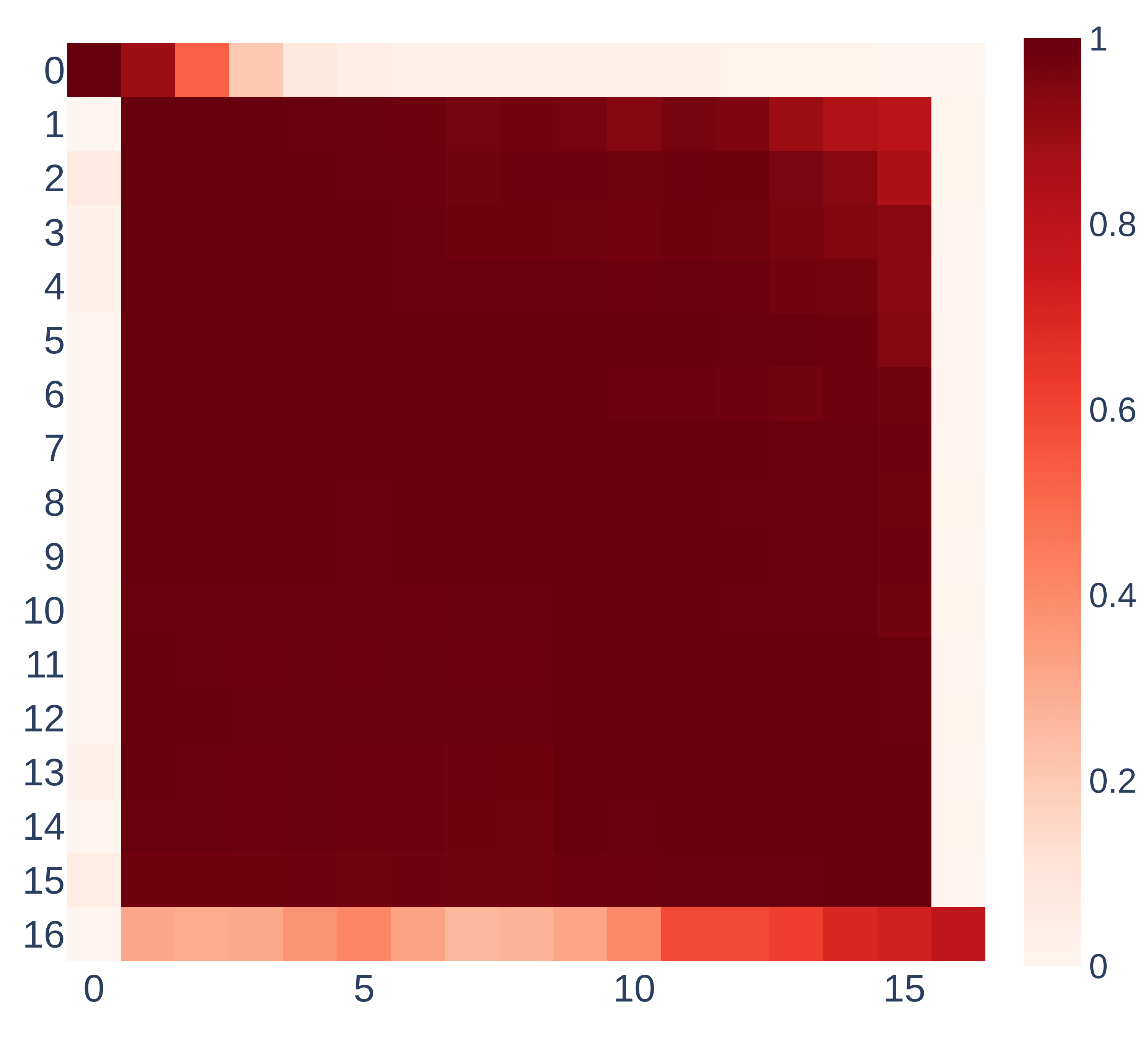}
    }
    \subfloat[\label{fig:all-v-one}probes trained on all-but-one layers, evaluated on the held-out layer]{
    \includegraphics[width=0.64\textwidth]{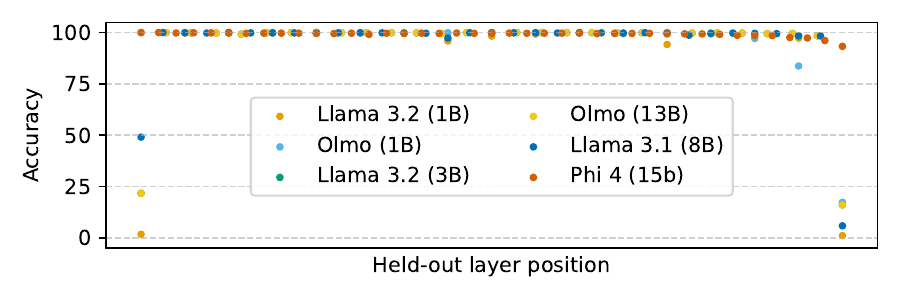}
    }
    \caption{\textbf{Transferability of representations}: Accuracy of param-sin probe evaluated on \textbf{unseen layers}
    \vspace{-5pt}
    }
    \label{fig:placeholder-x-layers}
\end{figure*}

Results of probes' cross-layer generalization for Llama 1B are displayed in \Cref{fig:placeholder-x-layers}. 
We observe that probes fitted on a concrete layer's representations (\Cref{fig:one-v-one}) generalize outstandingly well to both close-by and more distant layers, with the exception of the first and the last layer.
This trend reveals that language models \textbf{operate in a consistent representation} of numbers that is \textbf{universal throughout their computation}, undergoing only minor shifts across layers.
This trend is also corroborated by multi-layer probes evaluated in heldout-layer fashion (\Cref{fig:all-v-one}), reaching a close-to 100\% accuracy across \textit{all} intermediate layers and six models of diverse sizes and families.
This result has practical implications for future work in numeric interpretability --- showing that we can train universal, yet highly \textit{accurate} probes for intermediate layers of a broad set of language models.\makeatletter\ifacl@anonymize\footnote{Training scripts, together with reproducibility guidelines, will be provided upon acceptance.}\fi\makeatother

To summarize, we find that the same key, i.e. probe can be used to interpret numeric information at different layers of the same language model. Added to our previous observations in \Cref{sec:universal:inputs,sec:universal:contexts}, we can therefore stress that models learn input embeddings for numbers that are \textbf{strikingly similar}, regardless of which model they come from. These embeddings are processed into sinusoidal representations that are systematic and consistent regardless of context or hidden layer, with the source of this consistency maintained primarily by the residual stream across layers (cf. \Cref{appendix:tracking}).

\section{How to Study Numbers in LLMs}

We have established that models universally converge to sinusoidal number representations. This characteristic is not shared with other word pieces. This questions the appropriateness of existing interpretability tools when it comes to numbers and other ordinal data types. 

\begin{figure}[ht!]
    \centering
        \includegraphics[max width=0.85\columnwidth]{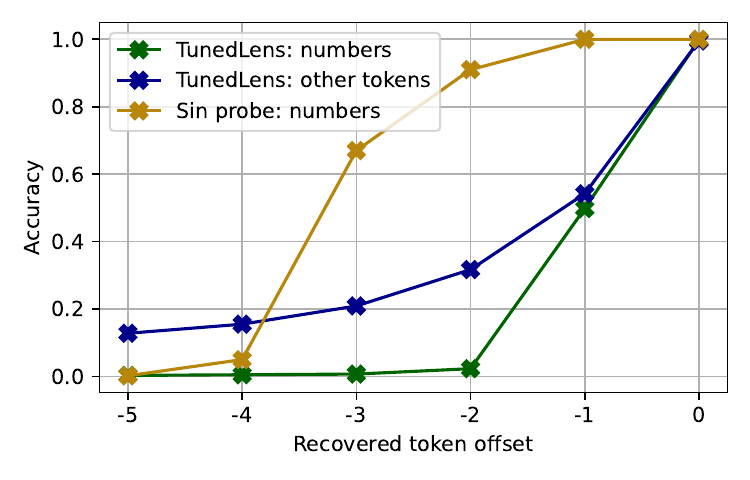}
        \vspace{-3pt}
        \caption{Accuracy of extracting pieces of \textbf{multi-token numbers} from the last token representations.
        \vspace{-6pt}
        }
        \label{fig:multitok-recovery}
\end{figure}

\subsection{Multi-token Numbers}
\label{sec:multitok}

As a case study for observing how the choice of interpretability toolset affects our findings, we analyse whether our conclusions about the remarkable accuracy of numeric representations also hold for large, \emph{multi-token} numbers.

First, we use our sinusoidal probes to retrieve the values of \textit{all parts} of multi-token numbers from a \textit{single}, i.e. last token's representation.
For this experiment, we adapt the methodology utilizing natural-language contexts (see \Cref{fig:nat lang}), with a substitution of original numeric values with values spanning two to six numeric tokens, i.e. in a numeric range between $10^3 - 10^{18}$.
We use a subset of BBC data that was published after the cutoff or release date of our LLMs \citep{Li_Guerin_Lin_2024}.
%

%
%
As a point of comparison, we consider the general-purpose TunedLens tool \citep{belrose2025elicitinglatentpredictionstransformers}, 
an early-exit interpretability method which aims to explain the contents of intermediate representations by learning to translate them into logits. 
For each layer $L_i$, TunedLens distills the computations done in the subsequent layers $L_{i+1}, \dots L_n$ into a simple affine transformation $\mathbf{W}_i$, such that applying this transformation $\mathbf{W}_i$ to a hidden state at $L_i$ followed by the unembedding matrix closely matches the logits that the LLM would eventually produce.
TunedLens was applied to number representations in previous work \citep{NEURIPS2024_2cc8dc30} and also represents the line of work arguing for a linear representations of numeric values in LLMs \citep{zhu-etal-2025-language}.
To maximize comparability and align the number and distribution of target categories, we narrow down the space of probed tokens to 1,000 tokens ranked as 2,000--3,000\textsuperscript{th} most-common tokens in our dataset (matching the number of classes and their distribution for our number probes).
Crucially, we also report previous-token recovery rates using TunedLens for \textit{number} tokens, using the same configuration as for natural-language tokens.


Results for Llama 1B and different probing methods are presented in \Cref{fig:multitok-recovery}.
Results of sin probes show that \textbf{multi-token numbers are indeed systematically and highly accurately \textit{superposed} in the latest numeric representation} -- reaching an accuracy of 99\% for the immediately-preceding number piece (offset -1). 
However, the accuracy and/or systematicity of the superposition mechanism quickly drops for numbers longer than \textit{three} tokens (i.e. $\geq 10^9$), reaching close to zero for the 5\textsuperscript{th} preceding token.

We notice that \textbf{TunedLens gives a conflicting account}.
While it shows that previous non-number tokens are also superposed, i.e. can be decoded as far as five time-steps later with non-zero probability, it suggests that this is not the case for number tokens -- these cannot be retrieved past two time-steps, and even then, only with 50\% accuracy.
However, a comparison with the sin probe shows that this trend is merely a result of TunedLens's inefficiency for numeric tokens; estimates derived from it provide an inaccurate representation of LLMs' accuracy, underestimating the amount of contained information and overlooking the superposition mechanism uncovered by a more suitable probe.




\begin{figure*}[th]
    \centering
    \setlength{\tabcolsep}{2pt}
    \renewcommand{\arraystretch}{1.0}

    \newcommand{\rowlabel}[1]{%
        \makebox[0pt][r]{%
            \raisebox{0.11\textwidth}[0pt][0pt]{%
                \rotatebox[origin=c]{90}{\footnotesize #1}\hspace{0.8em}%
            }%
        }%
    }

    \begin{tabular}{cccc}
        \vspace{-20pt}
        \rowlabel{Llama 3.2 3B}%
        \includegraphics[width=0.223\textwidth]{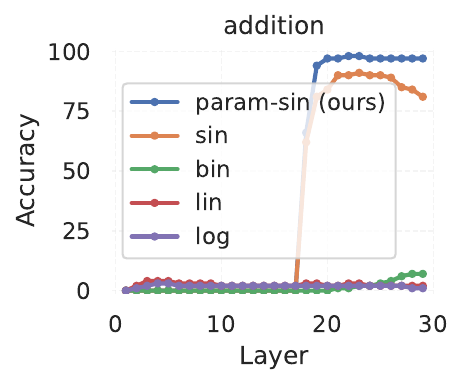} &
        \includegraphics[width=0.224\textwidth]{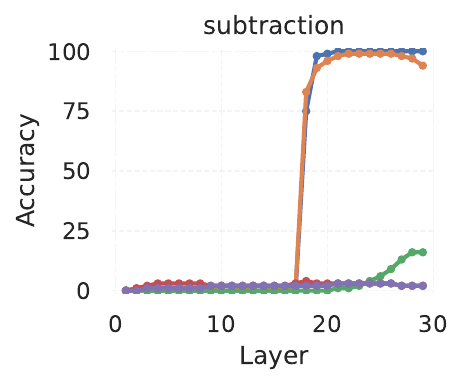} &
        \includegraphics[width=0.224\textwidth]{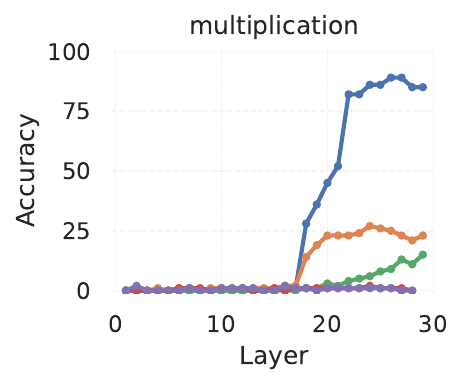} &
        \includegraphics[width=0.224\textwidth]{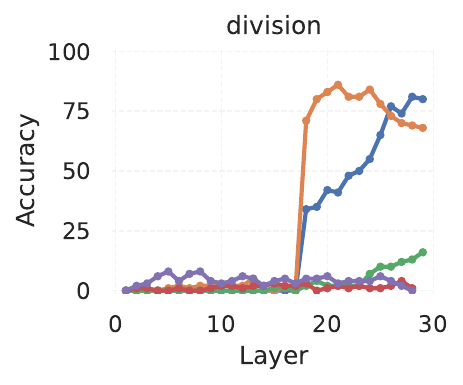} \\[1.2em]
        \vspace{-20pt}
        \rowlabel{OLMO 2 7B}%
        \includegraphics[width=0.224\textwidth]{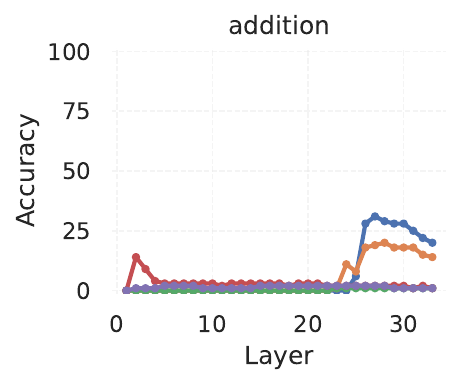} &
        \includegraphics[width=0.224\textwidth]{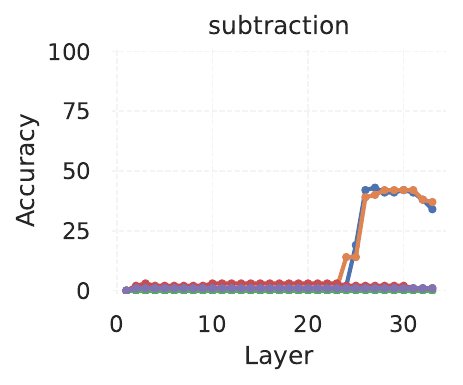} &
        \includegraphics[width=0.224\textwidth]{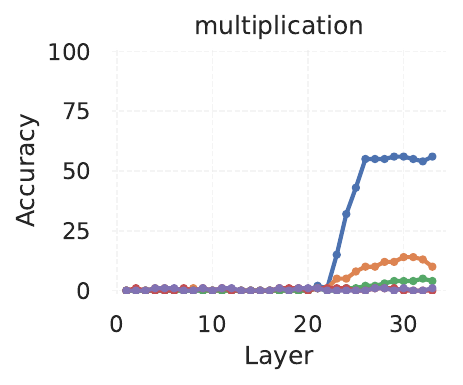} &
        \includegraphics[width=0.224\textwidth]{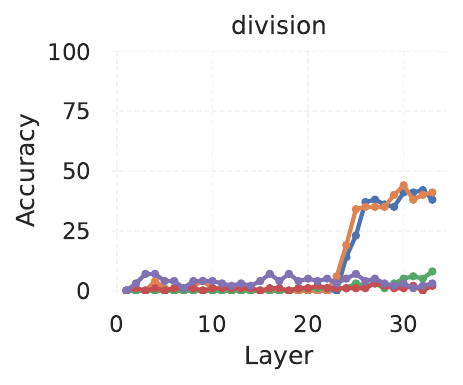} \\[1.2em]
        \vspace{-10pt}
        \rowlabel{Phi 4 15B}%
        \includegraphics[width=0.224\textwidth]{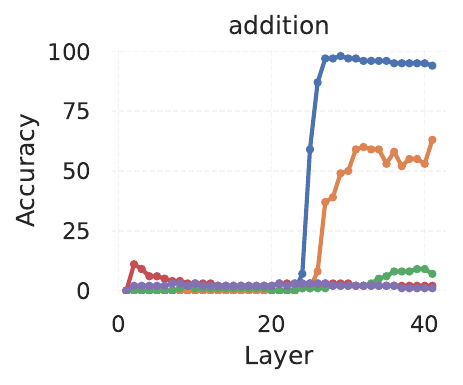} &
        \includegraphics[width=0.224\textwidth]{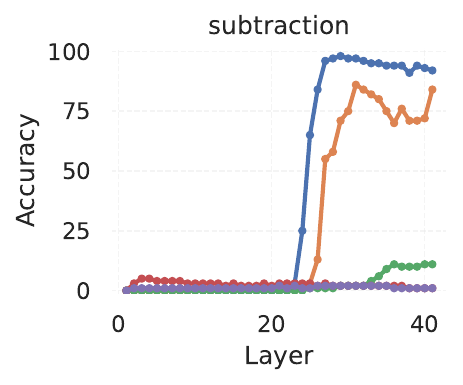} &
        \includegraphics[width=0.224\textwidth]{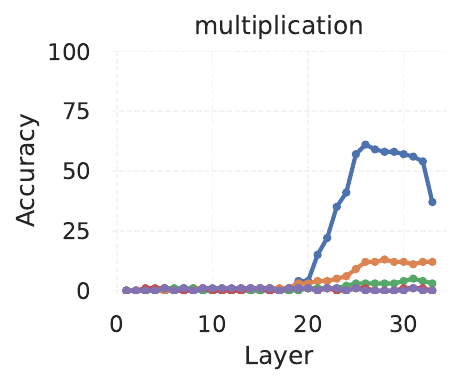} &
        \includegraphics[width=0.224\textwidth]{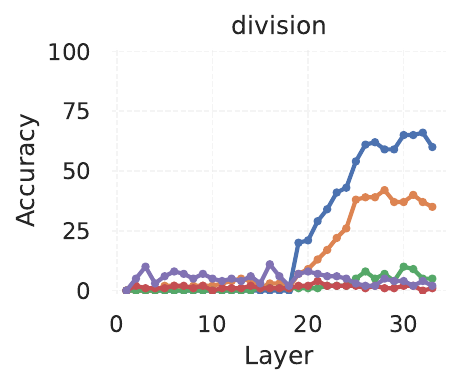}
    \end{tabular}

    \caption{\textbf{Probing accuracy} in extracting correct output from internal activations across layers; different probes.\vspace{-5pt}}
    \label{fig:output_results_across_probes}
\end{figure*}

\subsection{Output Tracking}
\label{sec:error-tracking}

A reliable mechanism of numeric manipulations necessitates an accurate representation of inputs (evidenced in §\ref{sec:universal}), but also the ability to construct systematic and accurate representations of \textit{outputs}.
Without the specialized tools to analyze numeric representations, prior work describes LLMs' arithmetical mechanisms through qualitative assessments limited to the outputs of addition \cite{kantamneni2025languagemodelsusetrigonometry,NEURIPS2024_2cc8dc30,levy-geva-2025-language}.
Therefore, having found the sinusoidal probes to be highly suitable for describing the representations of input numbers, we set off to investigate whether such probes can also better capture the mechanics of LLMs' \textit{output} representations, compared to other probe types.

In this set of experiments, we prompt the language models with prompts of the form ``\textit{The result of $x_1 \otimes x_2$ is }''
\footnote{Exact prompt formulations for each operation can be found in the supplementary material under \texttt{/notebooks/output\_probing}.}
~and train probes to predict the correct result of $x_1 \otimes x_2$, where $\otimes \in \{+, -, \times, /\}$.

Figure~\ref{fig:output_results_across_probes} shows the results for probes from prior work, including linear probes from \citet{zhu-etal-2025-language,NEURIPS2024_2cc8dc30} and a sin probe proposed and applied by \citet{kadlčík2025pretrainedlanguagemodelslearn}.
We can see that the param-sin probe performs comparable or better than probes applied for numbers in previous work and newly allows the tracking of outputs with over 80\% of accuracy in multiplication.
The results for addition also push the known lower bounds of precision of the result emerging early in the model, conflicting with a hypothesis of ``gradually refining predictions'' from \citet{NEURIPS2024_2cc8dc30} based on LogitLens, but supporting the interpretation of arithmetics implementation in LLMs by \citet{stolfo-etal-2023-mechanistic} through a combination in late layer(s).

\begin{figure}[th!]
    \centering
    \vspace{-6pt}
    \subfloat[Addition]{
    \includegraphics[width=0.2\textwidth,height=0.175\textwidth]{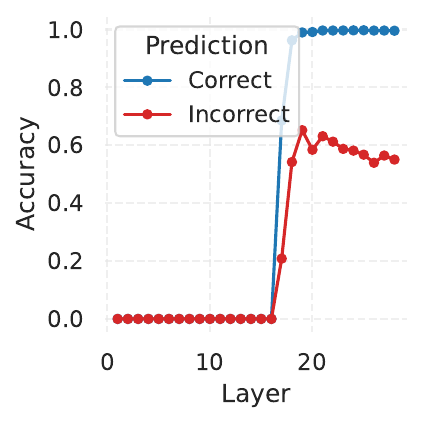}
    }
    \subfloat[Subtraction]{
    \includegraphics[width=0.2\textwidth,height=0.175\textwidth]{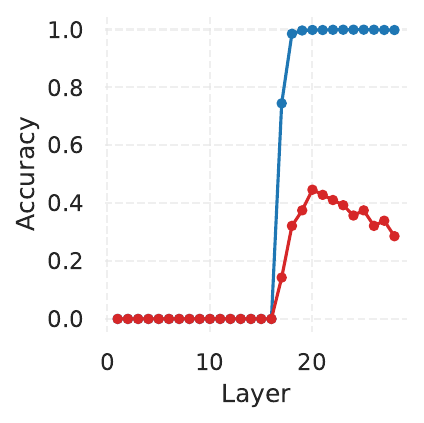}
    } 
    \vspace{-10pt}
    \subfloat[Multiplication]{
    \includegraphics[width=0.2\textwidth,height=0.175\textwidth]{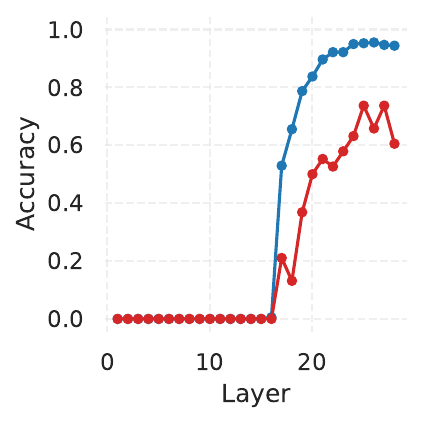}
    }
    \subfloat[Division]{
    \includegraphics[width=0.2\textwidth,height=0.175\textwidth]{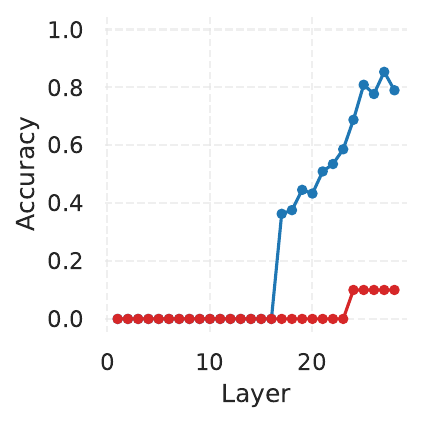}
    }
    \caption{\textbf{Probing accuracy} in extracting correct output from Llama 3.2 3B's internal activations separates model's correct (blue) and incorrect (red) predictions.}
    \label{fig:target-results}
\end{figure}

\Cref{fig:target-results} further details on a distinction between the cases where the model predicts a correct and incorrect response -- suggesting that the \textbf{extent to which the model maintains the sinusoidal representation may determine its accuracy}.

We evidence this by following qualitative and mechanistic evidence.
In \Cref{appx:output_embeddings}, we observe the output embeddings of addition and multiplication -- latter coupled with lower model's accuracy -- differ in the precision of the the observed sinusoidality; with the pattern being more prominent in the better-performing, addition embeddings.
In \Cref{sec:incr}, we further analyze how, according to our probes, is the result refined via gradually minimizing the response error across layers, with particular layers deteriorating this refinement, suggesting a worse fit of the sinusoidal probe and thus, the sinusoidal pattern.
We find that simply dropping the layers causing a divergence from the probe's fitted pattern lead to an accuracy improvement in multiplication and division in 4 out of 6 cases.

Finally, we evidence this point more directly by a mechanistic intervention -- \textbf{steering activations} in prediction cases leading to incorrect prediction \textbf{towards the expected sinusoidal pattern}. Concretely, we (i) fit sinusoidal probes at each layer to predict the output of a given arithmetic operation; (ii) optimize a set of 1,000 randomly-initialized `patch' embeddings $\mathbf{e}_{y}$ to maximize the probability that the probes will map them onto the integer value $y$; and (iii) steer activations towards the corresponding patch whenever the model is not producing the expected output. Steering is achieved by interpolating between the activation $\mathbf{h}_i$ at layer $L_i$ and the patch $\mathbf{e}_{y}$ for the target scalar $y$, $\alpha \mathbf{h}_i  + (1 - \alpha) \mathbf{e}_y$.
We report results with respect to the optimal value of $\alpha$ per layer, operation and probe.

\begin{figure}[th]
    \centering
    
    \subfloat[Llama 3B]{
    \includegraphics[width=0.95\linewidth]{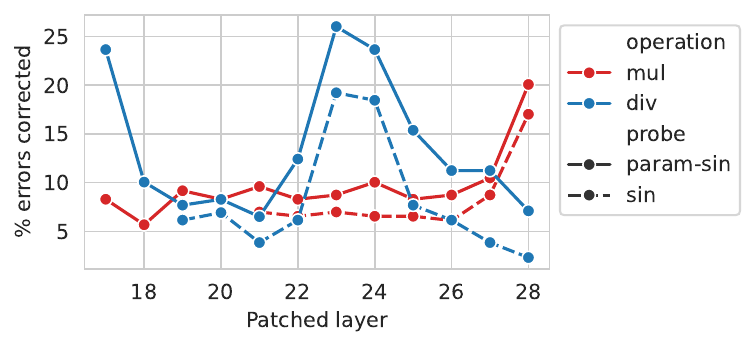}
    }
    \vspace{-8pt}
    \subfloat[OLMo-2 7B]{
    \includegraphics[width=0.95\linewidth]{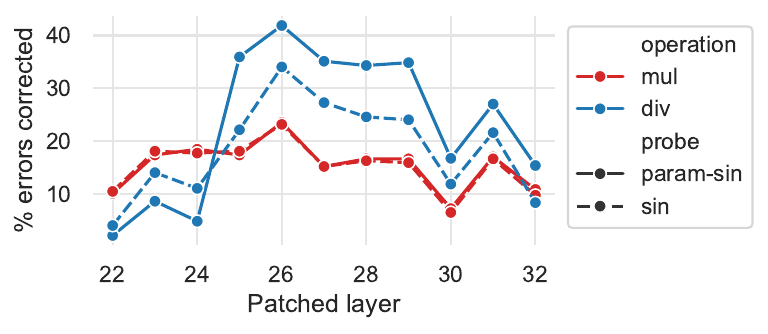}
    }
    \vspace{-8pt}
    \subfloat[Phi-4]{
    \includegraphics[width=0.95\linewidth]{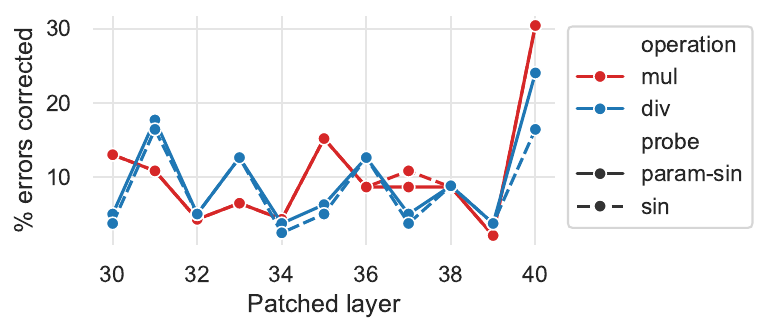}
    }
    \caption{\textbf{Error reduction} after \textbf{steering activations} on incorrect predictions of multiplication and division. Optimal interpolation factor $\alpha$ per setup.
    }
    \label{fig:steering}
\end{figure}

\Cref{fig:steering} displays the absolute difference in performance caused by the steering on previously erroneous cases in multiplication and division for three models. We target probes on upper layers which reach accuracy scores $\ge 10\%$. 
In short, depending on the layer, we can correct up to $30\%$ of errors for multiplication, and up to $42\%$ for division.
Crucially, our probe almost always outperform the probe proposed by \citet{kadlčík2025pretrainedlanguagemodelslearn}: Across all setups, we only observe four setups where the sin probe outperforms our sin-param probe.\footnote{Viz., division with OLMo at L22, L23, L24), multiplication with Phi at L37.} 
The optimal setup per model and operation is always obtained with our probe. 
This underscores once again the importance of selecting the right explanatory tool, this time in the context of activation steering.

In sum, factoring in the knowledge of the sinusoidal nature of number representations allowed us to understand and assess the lower bounds of accuracy of multi-token number representations. On a case study of Llama 3.2 3B, we further showed how more accurate probes allow pinpointing and \textbf{eliminating} sources of errors. 
Tools that are best suited to number representations allow us to enhance accuracy and perform more effective interventions.

\subsection{Ordinal Data Types}
\label{sec:ordinal}

Having established that sinusoidal mechanisms robustly encode numeric values and our specialized probe facilitates targeted interventions, we investigate how this tooling generalizes to ordinal data types -- potentially represented in the same systematic fashion as numbers.
A particular challenge here is a data scarcity -- for instance, as there are only seven days in a week, to warrant held-out evaluation, we can only fit the probes on six examples.

Prior work by \citet{engels2025not} demonstrates that LLMs internally represent temporal data -- such as days of the week and months of the year in circular structures. Analytically, a centered circle is parameterized by the equations $x(\theta) = r\sin(\theta + \pi/2)$ and $y(\theta)=r\sin(\theta)$, indicating that these temporal concepts are encoded using sinusoidal features. Assuming this structure, we fit our probe on the embedding representations of these concepts and compare it to a prior probe architecture of \citet{kadlčík2025pretrainedlanguagemodelslearn} in leave-one-out cross-validation setting. The best held-out validation accuracies of both architectures are reported in~\Cref{tab:probing-ordinals}.
\begin{table}[tbh]
\centering
\small
\begin{tabular}{lcc}
\toprule
\textbf{Model} & \textbf{sin} & \textbf{param-sin} \\
\midrule
\multicolumn{3}{c}{\textit{Days of week (n=7)}} \\
\midrule
Olmo 2 1B & 4 & \textbf{6} \\
Olmo 2 7B & 6 & 6 \\
Olmo 2 13B & 5 & \textbf{6} \\
Llama 3 1B & 5 & \textbf{6} \\
Llama 3 3B & \textbf{7} & 6 \\
Llama 3 8B & 3 & \textbf{6} \\
\midrule
\multicolumn{3}{c}{\textit{Months (n=12)}} \\
\midrule
Olmo 2 1B & 6 & \textbf{11} \\
Olmo 2 7B & 6 & \textbf{11} \\
Olmo 2 13B & 5 & \textbf{11} \\
Llama 3 1B & 9 & \textbf{11} \\
Llama 3 3B & 10 & \textbf{11} \\
Llama 3 8B & 6 & \textbf{11} \\
\midrule
\multicolumn{3}{c}{\textit{Numerals: `one' to `ten' (n=10)}} \\
\midrule
Olmo 2 1B & 3 & \textbf{6} \\
Olmo 2 7B & 4 & \textbf{8} \\
Olmo 2 13B & 4 & \textbf{8} \\
Llama 3 1B & 3 & \textbf{5} \\
Llama 3 3B & 4 & \textbf{6} \\
Llama 3 8B & 3 & \textbf{9} \\
\bottomrule
\end{tabular}
\caption{Number of matches achieved by training param-sin and best-performing prior probe (sin) in decoding ordinal data with extremely small training sets from models' internals. Held-one-out evaluations.}
\label{tab:probing-ordinals}
\end{table}
Our results demonstrate high decoding accuracy for ordinal data across all evaluated models. This validates the probe's utility beyond pure numbers, even under extreme data-scarcity. With our better generalizing architecture, we corroborate and extend the discovery by~\citet{engels2025not} of circularity in LLMs' representations to other ordinal data types.

\section{Conclusions}

In this paper, we have shown the number representations in LLMs are exceptional in two distinct but related ways.
First, the regularity and systematicity of the sinusoidal structure that all LLMs converge is striking, to the point of being almost universal across all experimental setups we consider. We can view the same frequencies in Fourier being used by all models we survey (§\ref{sec:universal:inputs}), we can train a probe on one layer and apply to another (§\ref{sec:universal:x-layers}). The one remarkable exception is the discrepancy between natural language contexts, which yield remarkably robust probes that generalize broadly, and synthetic probes derived from templated texts, which do not.

Second, this exceptional sinusoidal regularity requires adapting the way we study number processing in LLMs. Using a general-purpose TunedLens method, applied for numbers in prior work, grossly underestimates how accurately previous timestep information can be retrieved, whereas sinusoidal probes such as the one we propose in \cref{eq:new-sin} may lead to substantially higher accuracies and contradictory conclusions (§\ref{sec:multitok}).
Finally, we demonstrate that the sinusoidal probes are also well applicable for analyzing the emergence and manipulation of model's own predictions and that the accuracy of representation in terms of sinusoidality is related to the prediction accuracy (§\ref{sec:error-tracking}).

While we focus on numbers, our experiments showcase how important the appropriate interpretability toolset is for accurate interpretation of specialized data types.
This naturally questions the common practice of designing \textit{general-purpose} interpretability and explainability tools, motivating future work towards adapting our practices to the specifics of the type of data we want to explain.


\section*{Limitations}

Several practical factors limit the scope of our work. 
We focus our survey on three particular model families; moreover, we have excluded very large models (>30B parameters) due to practical hardware limitations. 
The rapid pace of modern NLP may entail that the set of models we have picked for this study will no longer be representative of contemporary frontier models in the future.

Our work is also based on English-centric LLMs, but the number systems vary greatly across languages --- e.g., Chinese and Japanese often employ myriad-based number systems and frequently use both Chinese and Arabic numerals. The extent to which our conclusions will hold for a Chinese-centric or Japanese-centric LLM remains a topic for future research.

Lastly, we find that all errors in probing temporal concepts (days, months) are exactly in the first position (Monday, January), persisting even when the probed token list is reversed (Sunday, December). We suspect that zero value is a blind spot of the probe due to its parametrization, which needs to be addressed to achieve a perfect decoding rate.


\makeatletter
\ifacl@finalcopy
\section*{Acknowledgments}

This work is supported by the Research Council of Finland through project No.~353164 ``Green NLP -- controlling the carbon footprint in sustainable language technology''.

\noindent \euflag \ This project has received funding from the European Union’s (EU) Horizon Europe research and innovation programme under Grant agreement No. 101070350. 
The work has also received funding from the Digital Europe Programme under grant agreement No 101195233 (OpenEuroLLM). 
The contents of this publication are the sole responsibility of its authors and do not necessarily reflect the opinion of the EU.

\fi
\makeatother

\bibliography{custom}

\appendix

\section{Supplementary results}

\subsection{Selecting the optimal number of Fourier components for comparison}
\label{appendix:fft threshold}

\begin{figure}[ht]
    \centering
    \includegraphics[max width=\linewidth]{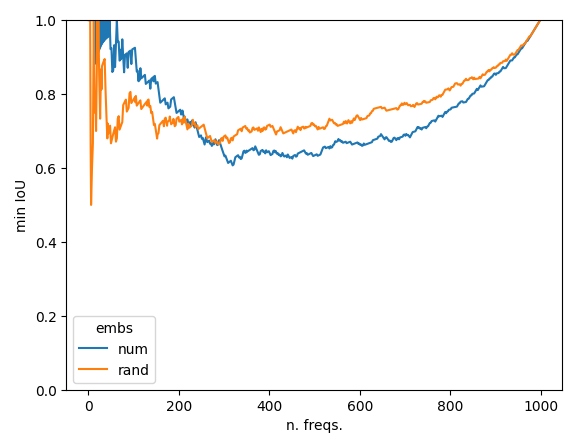}
    \caption{Minimum IoU across number of frequencies considered}
    \label{fig:iou by k}
\end{figure}

In \Cref{sec:universal:inputs}, we discuss selecting the optimal number of Fourier components.
As can be assessed from \autoref{fig:fft iou:example}, from a certain frequency rank, Fourier decompositions resort to noise.
Thus, our objective is to find the ``cutoff'' rank ($k$) that disentangles oscillating information from the non-oscillating, presumably non-numeric information. This means finding an \textit{optimal}, yet non-trivial (e.g. $\geq 10$) cutoff for a number of components.
In practice, we evaluate for each value of $k$ what the minimum IoU agreement score across all models amounts to for numbers and random overlapping word-pieces.
As displayed in \Cref{fig:iou by k}, we find such an optimal cutoff at $k=63$ as the highest value of $k$ that leads to perfect agreement across models. 

\begin{figure}[ht!]
    \centering
    \includegraphics[max width=\linewidth]{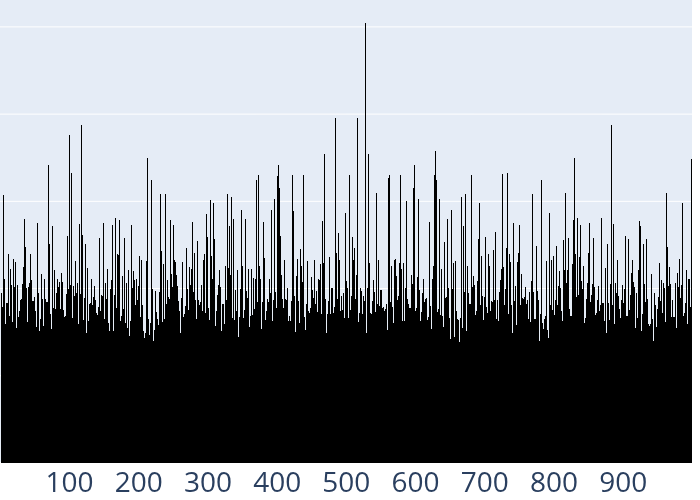}
    \caption{Contributions of Fourier base frequencies for non-number pieces (Llama-3 1B, provided as an example)}
    \label{fig:fft rand}
\end{figure}

A more thorough assessment of \Cref{fig:iou by k} suggests a few interesting trends. Random word pieces also tend to favor a handful of Fourier basis components; which we conjecture is due to the sampling mechanism. By selecting overlapping pieces, our pieces must be frequent enough to be present in multiple distinct tokenizers, which in turns shapes the type of linguistic units represented in this random sample.
Secondly, we also observe a `cross-over' point around $k\approx 250$, after which we find greater agreement in random pieces than numbers. Yet, we note that (i) a  significant proportion of the mass is concentrated in a few frequency components for numbers, whereas random pieces lead to much more uniform distributions across frequency components (see \Cref{fig:fft rand}); and (ii) the results in \Cref{fig:iou by k} still allow us to establish that the Fourier profile of number pieces is clearly distinct from what we observe for any other overlapping set of word pieces.

\subsection{Natural-language contexts for representations extraction}
\label{appx:natural_lang_contexts}

Contrary to prior work, we use the contexts of natural language to extract the representations of numbers. We argue that such setting better corresponds to the practical applicability of probes -- in monitoring models' behavior and errors, or in tracking the faithfulness of reasoning chains.

We choose the contexts with medium to high co-dependency on numerical contents covering four distinct domains and eight different datasets.
The datasets categorized into these domains are presented in \autoref{tab:lan-context-datasets} and total in around 700 thousands natural-language contexts.

\subsection{Generalization of probes to unseen layers}
\label{appendix:cross_layer_gen}

\begin{figure*}[ht!]
    \centering
    \subfloat[Llama 3 1B]{
    \includegraphics[width=0.4\textwidth]{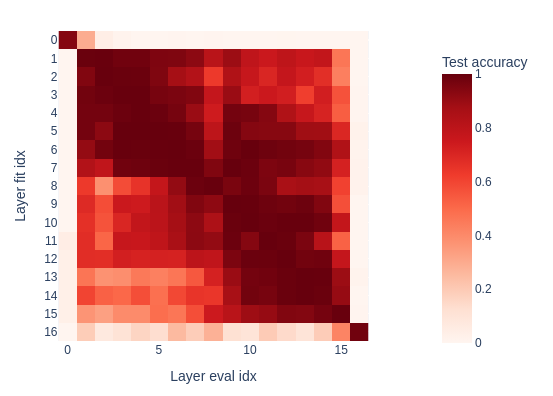}
    }
    \subfloat[Llama 3 3B]{
    \includegraphics[width=0.4\textwidth]{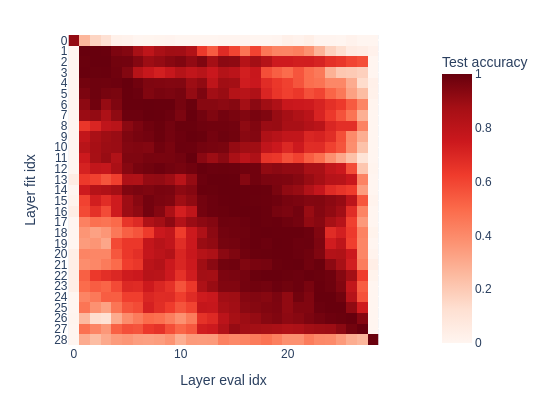}
    }
    \newline
    \subfloat[Llama 3 8B]{
    \includegraphics[width=0.4\textwidth]{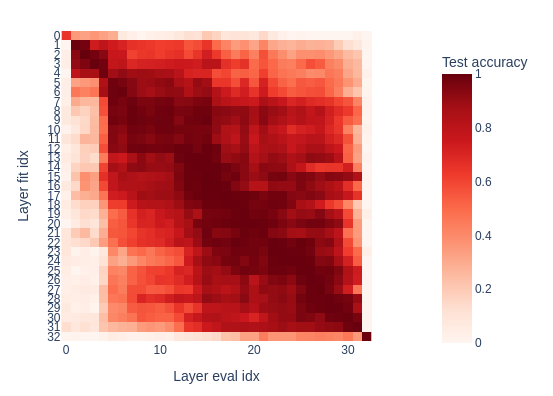}
    }
    \subfloat[Olmo 2 1B]{
    \includegraphics[width=0.4\textwidth]{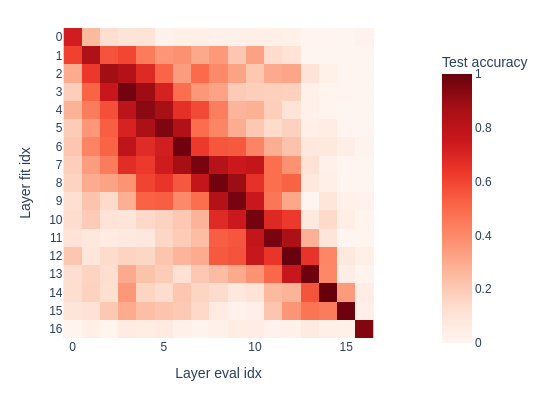}
    }
    \newline
    \subfloat[Olmo 2 7B]{
    \includegraphics[width=0.4\textwidth]{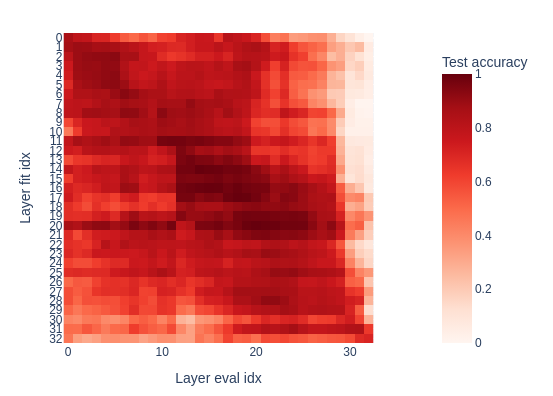}
    }
    \subfloat[Olmo 2 13B]{
    \includegraphics[width=0.4\textwidth]{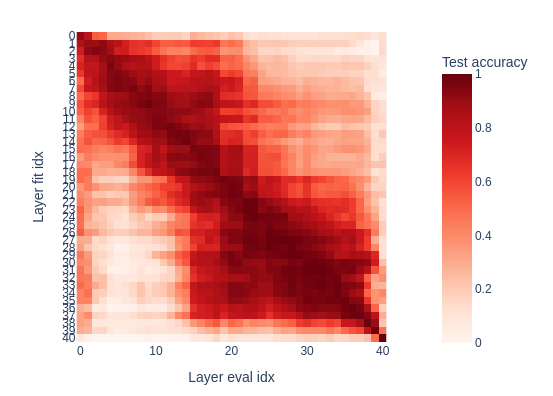}
    }
    \newline
    \subfloat[Phi 4 15B]{
    \includegraphics[width=0.4\textwidth]{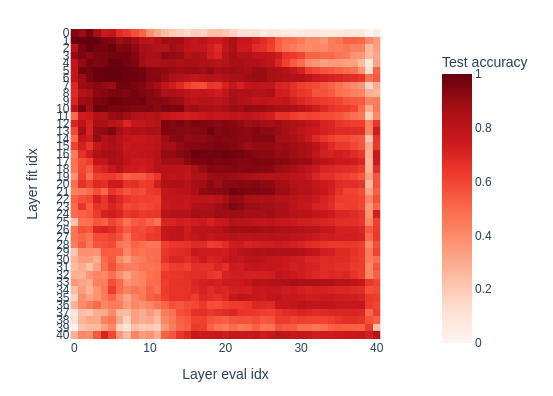}
    }

    \caption{Probe fitted on one layer evaluated on all layers, synthetic math contexts. Olmo 2 1B shows the weakest cross-layer generalization among all models. Llama models display a strong separation between input/output embedding representations and the hidden representations. }
    \label{fig:crosslayer_generalization_all}
\end{figure*}

\Cref{fig:crosslayer_generalization_all} shows how probes fitted on each layer generalize to all other layers in a model, using synthetic mathematical contexts. Strong cross-layer generalization indicates a high consistency of representations throughout the model.

\subsection{Visualizations of internal activations}
\label{appendix:pca_fourier}

We visualize the internal activations for a string template \enquote{x1 + x2 =} on the second numeric token (x2). Values for x2 are selected as a range 0-999 (all values present by the vocab), and values for x1 are sampled randomly from the same range. Then, we project the activations of the model's middle layer to 64 dimensions with PCA and compute the Fourier transform.

\begin{figure*}
    \centering
    \subfloat[Llama 3 1B (layer 8/17)]{
        \includegraphics[width=0.95\linewidth]{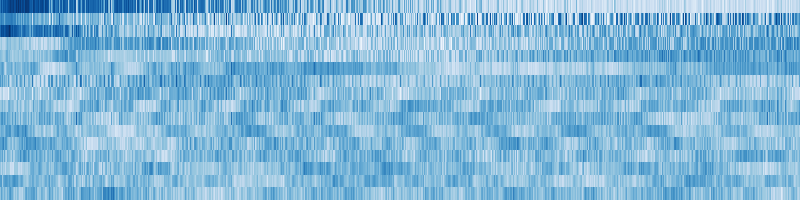}
    }
    \newline
    \subfloat[Olmo 2 7B (layer 16/31)]{
        \includegraphics[width=0.95\linewidth]{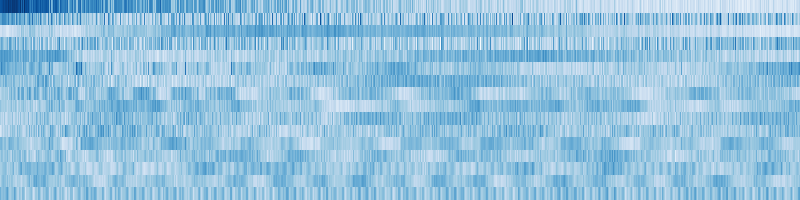}
    }
    \newline
    \subfloat[Phi 4 15B (layer 20/41)]{
        \includegraphics[width=0.95\linewidth]{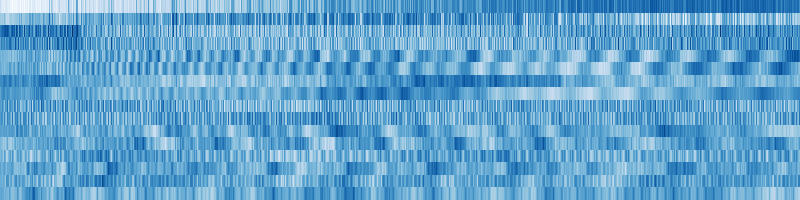}
    }
    \caption{PCA of models' internal representations}
    \label{fig:pca_hidden_states}
\end{figure*}

\begin{figure*}
    \centering
    \subfloat[Llama 3 1B (layer 8/17)]{
        \includegraphics[width=0.95\linewidth]{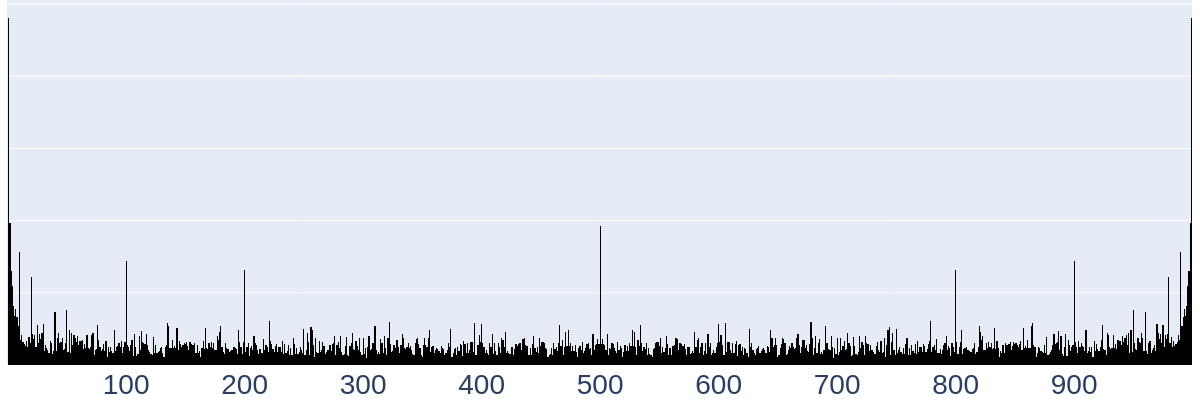}
    }
    \newline
    \subfloat[Olmo 2 7B (layer 16/31)]{
        \includegraphics[width=0.95\linewidth]{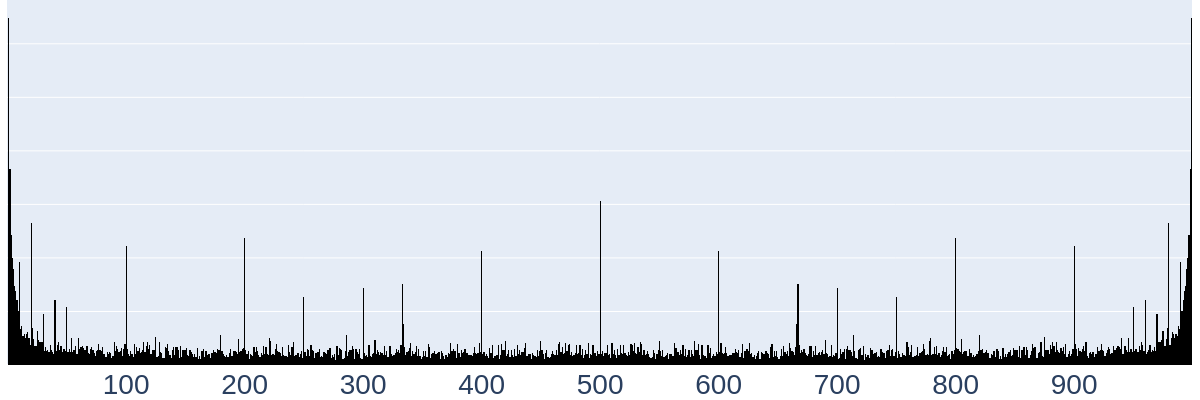}
    }
    \newline
    \subfloat[Phi 4 15B (layer 20/41)]{
        \includegraphics[width=0.95\linewidth]{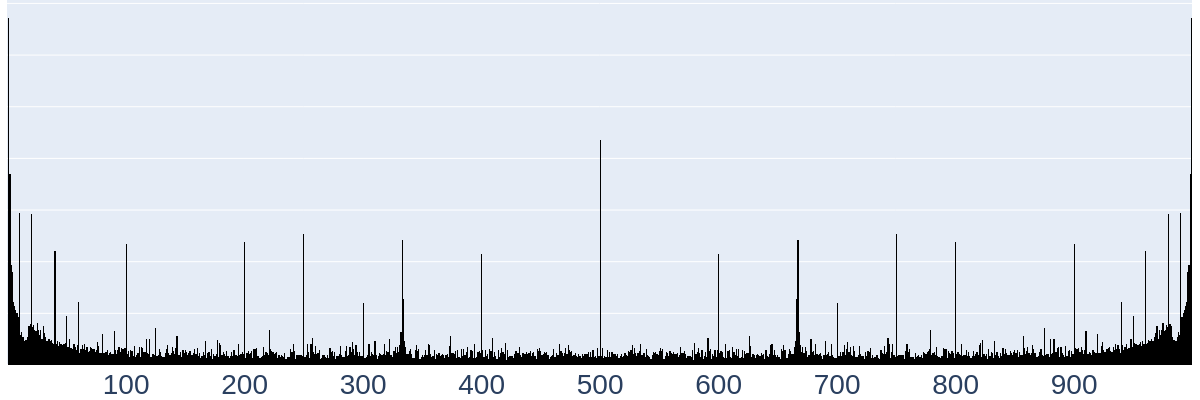}
    }
    \caption{Maximal magnitudes of frequencies in Fourier transform of PCA of models' internal representations}
    \label{fig:fft_hidden_states}
\end{figure*}

We visualize the first 16 PCA dimensions in \Cref{fig:pca_hidden_states} and the maximal magnitudes of the frequencies in the Fourier transform in \Cref{fig:fft_hidden_states}. We visualize across model sizes and families.

\begin{table}
\centering
\resizebox{0.95\columnwidth}{!}{
\begin{tabular}{lllll}
\toprule
 & Add & Sub & Mul & Div \\ \midrule
Accuracy & 100\% & 99.8\% & 90.4\% & 5.9\% \\
P(Extracted $|$ Incorrect) & - & 56.8\% & 26.3\% & 94.4\% \\
P(Not extracted $|$ Correct) & 1.4\% & 0.03\% & 26.1\% & 5.9\% \\ \hline
\end{tabular}
}

\caption{\textbf{Probe accuracy in extracting the predicted result:} Ratio of cases where the sin probe of some layer (top) retrieves a correct result when the model's prediction is \textit{incorrect}, and (bottom) can \textit{not} retrieve a correct result when the model's prediction is \textit{correct}.}
\label{Tab:probe_extraction_acc}
\end{table}

\subsection{Qualitative assessment of output embeddings}
\label{appx:output_embeddings}

\label{appendix:multiplication_different_repr}

\begin{figure}
    \centering
    \subfloat[addition]{
        \includegraphics[width=0.95\linewidth]{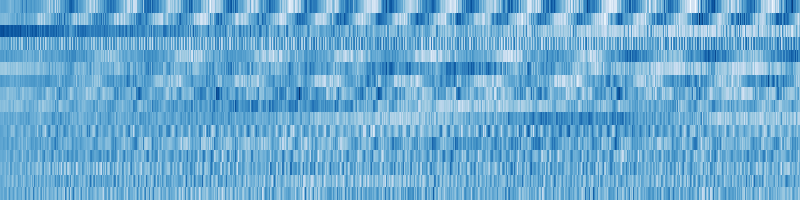}
    }
    \newline
    \subfloat[multiplication]{
        \includegraphics[width=0.95\linewidth]{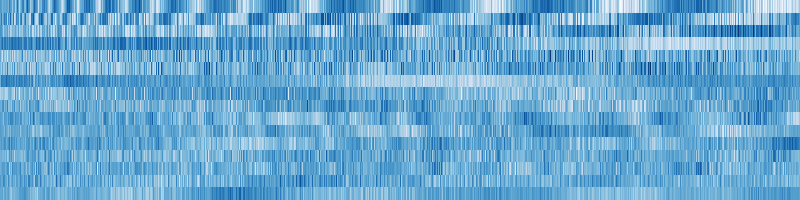}
    }
    \caption{PCA of Llama 3 1B \textit{output} representations for the operations of addition (recoverable with close-to 100\% accuracy) and multiplication (recoverable with appx. 80\% of accuracy)}
    \label{fig:llama_output_repr_pca}
\end{figure}

We visualize output representations of Llama 3 1B on addition and multiplication operations. For each expected result value $y$ in the range 0-999, we sample a random pair ($x1$, $x2$) from the same range, such that $x1+x2=y$ (or $x1\times x2=y$, respectively). We then predict the next token for prompts \enquote{{$x1$} plus {$x2$} is } and \enquote{{$x1$} multiplied by {$x2$} is } and collect the final output representations of the model before decoding. We then reduce the representations with PCA to 16 dimensions and visualize the result in \Cref{fig:llama_output_repr_pca}.

\subsection{Probe sparsity}
\label{appendix:sparsity}

Cross-layer evaluation indicates that models' internal representations tend to differ from the input/output embeddings. We find a notable discrepancy in the \textit{sparsity} of their representation; 
Whereas the input/output embeddings represent numbers in a more \textit{distributed} fashion, hidden layers use a small number of consistently-ordered sin features.


\begin{figure*}[ht!]
    \centering
    \includegraphics[width=\textwidth]{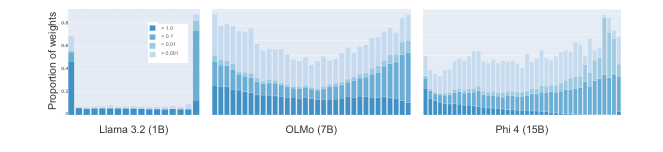}
    \caption{\textbf{Difference in probe weights distribution (of $W_{\text{in}}$) across layers} shows that the initial and last layers represent numbers in a systematically sparser fashion.}
    \label{fig:sparsity}
\end{figure*}

This is visualized in \Cref{fig:sparsity}, showing that intermediate layers in Llama 1B lead to probes with very few weights with values $> 10^{-5}$. 

We hypothesize that this discrepancy can be caused by the models' computational and optimization mechanics. As cross-entropy loss is minimized by confident one-hot predictions, models can benefit from using more features to create higher dissimilarity (separability) of tokens. In contrast, internal representations are not subject to direct optimization pressure and might benefit from fewer but more informative features.

Nevertheless, we note that this trend is pertinent across different models to a different extent: OLMo 7B and Phi 15B learn more distributed features also across the intermediate layers --- which, in turn, also leads to a better generalization of universal probes (\autoref{fig:placeholder-x-layers}b), but still causing an outstanding drop in held-out accuracy compared to other layers.
Finally, we note that the models' sparsity profile does not relate to reported performances on arithmetic tasks \cite{kadlčík2025pretrainedlanguagemodelslearn} and it also does not necessarily determine the accuracy of probes trained specifically for particular layer(s) (\Cref{fig:perf-l2l}).

\subsection{Using probes for tracking representation consistency}
\label{appendix:tracking}

\begin{figure}
    \centering
    \includegraphics[width=\columnwidth]{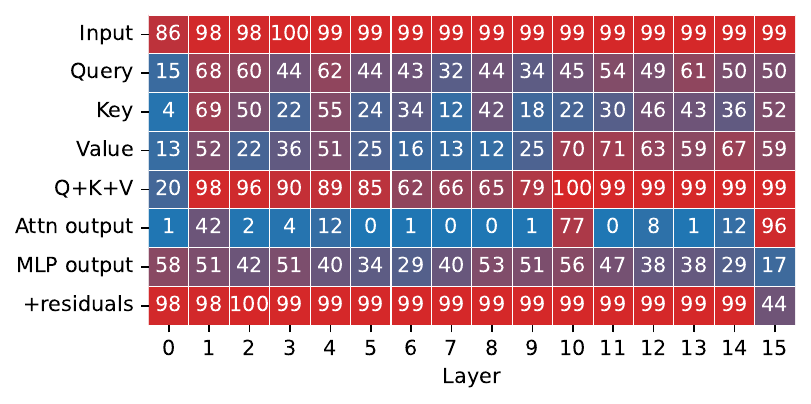}
    \caption{Accuracy of extracting input numbers from each component of the Transformer block using the sinusoidal probe.}
    \label{fig:in_layer_accs}
\end{figure}

We explore the origin of the high consistency of representations across layers within the internal mechanism that each transformer layer implements. 
In \autoref{fig:in_layer_accs}, we visualize the accuracy of probing input representations from each of the components present in transformer layers (Llama 3.2 1B for brevity).
We find the accurate numeric representation is scattered across different components in the attention mechanism, with the attention output projection largely violating the sinusoidal representation, which is then, to a large extent, reconstructed in the subsequent fully-connected block.
Nevertheless, the consistency of layers' output representation is primarily maintained by residual streams across layers. 

\subsection{Incremental refinement of model predictions on arithmetic tasks}
\label{sec:incr}

\begin{figure}[th!]
\vspace{-6pt}
    \centering
    \subfloat[Addition]{
    \includegraphics[width=0.23\textwidth]{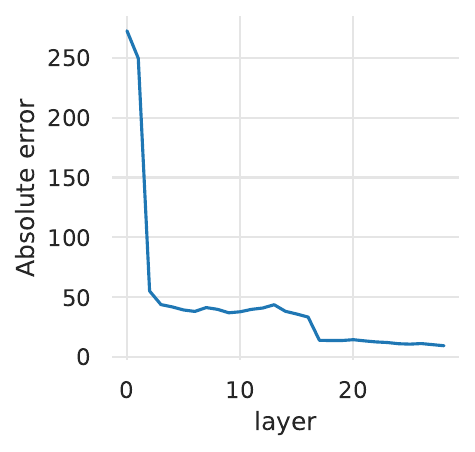}
    }
    \subfloat[Subtraction]{
    \includegraphics[width=0.23\textwidth]{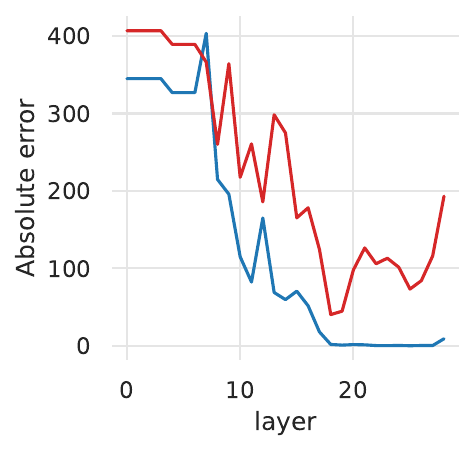}
    } 
    
    \subfloat[Multiplication]{
    \includegraphics[width=0.23\textwidth]{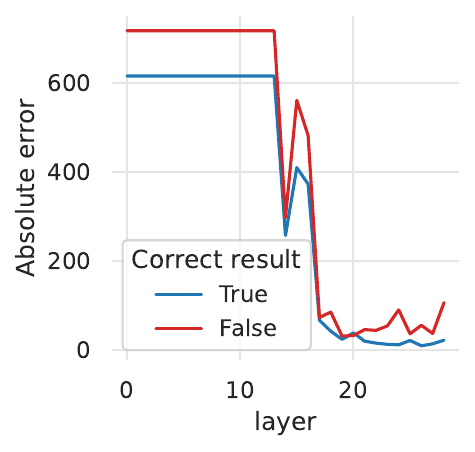}
    }
    \subfloat[Division]{
    \includegraphics[width=0.23\textwidth]{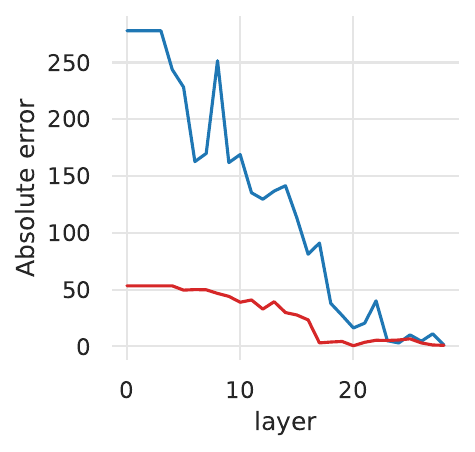}
    }

    \caption{\textbf{Absolute error per layer:} Absolute error of numeric values probed from different layers.}
    \label{fig:abs_error_per_layer}
\end{figure}

We can use sinusoidal probes to estimate the absolute error per layer: the absolute difference between the predicted and target integer provides an estimate of how far off from the intended target a model is at any intermediate step of computation. 
Results aggregated per layer of Llama 3.2 3B are displayed in \Cref{fig:abs_error_per_layer}. 
Across all operations, models tend to incrementally reduce the error towards the true answer value, \textit{gradually} refining across layers.
These results also suggest that there are particular layers responsible for an increase of errors in the model's internal computation.


    


\begin{figure}[th!]
    \centering
    \subfloat[Addition]{
    \includegraphics[width=0.23\textwidth]{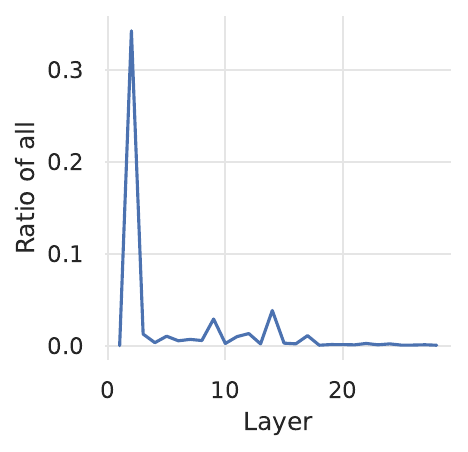}
    }
    \subfloat[Subtraction]{
    \includegraphics[width=0.23\textwidth]{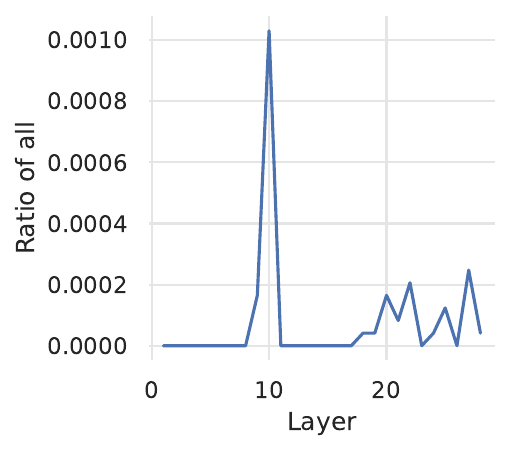}
    } 
    
    \subfloat[Multiplication]{
    \includegraphics[width=0.23\textwidth]{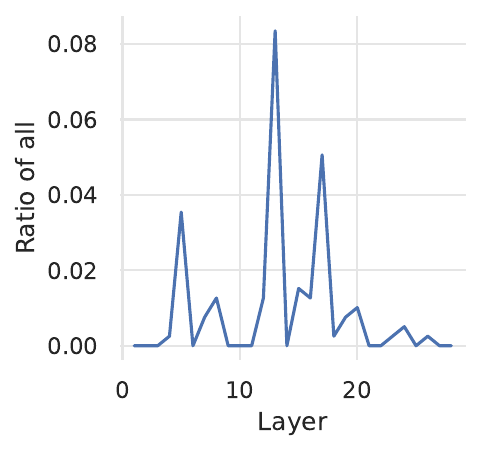}
    }
    \subfloat[Division]{
    \includegraphics[width=0.23\textwidth]{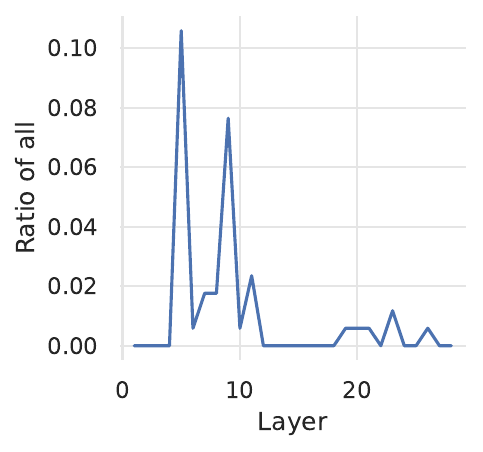}
    }

    \caption{\textbf{Error aggregation per layer:} Relative ratio of cases where each of the layers in Llama 3.2 3B \textit{breaks} the correct result from the previous layer. } 
    \label{fig:new_error_per_layer}
\end{figure}

\Cref{fig:new_error_per_layer} corroborates this hypothesis --- showing that particular layers `break' the correct result probed from the previous layer in large proportions of all predictions.
For instance, in division, the 5th layer is responsible for breaking the correct result recoverable from the previous layer in over 10\% of all cases.
Further analyses, presented in \Cref{Tab:probe_extraction_acc}, show that according to the probed representations, the models often achieve a correct result \textit{internally}, even though it does \textit{not} surface: this is the case of 56.8\% of all surfaced (prediction) errors in subtraction, 26.3\% in multiplication, and as much as 94.4\% of all errors in division.

Having used our probes to identify key layers that are likely responsible for errors, we can now turn to practical \textbf{interventions} that they make possible.
In practice, we simply remove the suspected layers from the model. This extremely naive intervention strategy is justified by the universality of representations across layers (\Cref{sec:universal:x-layers}).
We remove one of the three layers with the highest error aggregation, separately in multiplication and in division (having a potential for improvement of accuracy). We find that this enhances performance in \textit{four} out of six cases --- in one case (layer 4) in multiplication, causing a 26\% error reduction (from 90.38\% to 92.91\% and in \textit{all} cases in division, with error reduction between 27--64\%.
However, we note that the absolute number of correct predictions probed in the lower layers is negligible, which hinders the general applicability of this methodology.

\section{Experimental details}
\label{appendix:datasets}

The overview provided here is intended for informative purposes. For precise replication, we refer the reader to the companion codebase, which contains the exact hyperparameters for each experiment, available at \url{https://github.com/prompteus/numllama}.

In our probe architecture \textit{param-sin}, we initialize amplitude parameters $\mathbf{w}_a$ with a constant of $0.01$, phase-shift parameters $\mathbf{w}_\phi$ randomly from uniform distribution between $0$ and $2\pi$, and frequency parameters $\mathbf{w}_f$ with a vector of values regularly-spaced between $0$ and $\frac{\pi}{4}$, and the parameters of linear projection $\mathbf{W}_\mathrm{in}$ randomly from uniform distribution between $-\sqrt{1/n}$ and $\sqrt{1/n}$, where $d$ is the input dimension (number of model features).

We find that L1-regularization of the amplitude parameters $\mathbf{w}_a$ improves generalization to held-out tokens by encouraging the probe to maintain small effective latent dimensionality even with a large actual dimension $d$. Similarly, we found similar benefits from regularization of the linear projection $\mathbf{W}_\mathrm{in}$, although L2-regularization appears more robust. We find that an analogous configuration is also effective for the probe of \citeauthor{kadlčík2025pretrainedlanguagemodelslearn}, with L2-regularization of $\mathbf{W}_\mathrm{in}$ and L1-regularization of $\mathbf{W}_\mathrm{out}$ for inducing simplicity (sparsity) in the sinusoidal basis.

 The probes are fitted on a train set, with a checkpoint selected on the validation set, and then evaluated on the held-out test set. In most experiments, we use the Adam optimizer with a learning rate of $10^{-4}$ and apply regularization with a coefficient of either $10^{-3}$ or $10^{-2}$.

In multi-token decoding (\Cref{fig:multitok-recovery}), we use a learning rate of $5 \cdot 10^{-4}$, 10,000 training steps, and early-stop training.

When fitting probes on natural language contexts (\Cref{fig:nat lang}, `Nat. lang. $\rightarrow$ Nat. lang.') and for cross-layer transfer (\Cref{fig:crosslayer_generalization_all}), we use a hidden dimension of $d=500$.


In our steering experiments (\Cref{fig:steering}), for $f_{\sin}$, we optimize embeddings with respect to the probe using an SGD optimizer with a learning rate of $0.1$, optimize the embeddings for 200,000 steps, and decrease the learning rate by a factor of 100 over the first 100,000 steps.
As for $f_{\mathrm{param}\sin}$, we set regularization coefficient to $10^{-4}$ and use a hidden dimension of $d=2000$.

\begin{table*}[ht]
\centering
\resizebox{\textwidth}{!}{%
\begin{tabular}{@{}llllc@{}}
\toprule
\textbf{Domain} & \textbf{Dataset} & \textbf{Source} & \textbf{Numerical Context} & \textbf{Size} \\
\midrule
\multirow{2}{*}{Culinary} 
& Recipe NLG Lite & \texttt{m3hrdadfi/recipe\_nlg\_lite} & Quantities, measurements & 6118 \\
& FoodRecipe-ImageCaptioning & \href{https://github.com/samsatp/FoodRecipe-ImageCaptioning/blob/main/data/data_strings_local.json}{samsatp/FoodRecipe-ImageCaptioning} & Ingredient amounts & 719 \\
\midrule
\multirow{1}{*}{Temporal} 
& TimeLineExtraction & \href{https://github.com/irlabamsterdam/TimeLineExtractionDecisionLettersCASE/tree/main}{irlabamsterdam/TimeLineExtraction...CASE}& Legal document dates & 50 \\
\midrule
\multirow{3}{*}{Arithmetic} 
& MetaMathQA & \texttt{meta-math/MetaMathQA} & Mathematical reasoning & 395K \\
& DROP & \texttt{ucinlp/drop} & Discrete reasoning & 77.4K \\
& AQuA-RAT & \texttt{deepmind/aqua\_rat} & Algebraic word problems & 97.4K \\
\midrule
\multirow{2}{*}{Medical} 
& ICD-10 Codes & \texttt{atta00/icd10-codes} & Diagnostic codes & 25.7K \\
& ICD-10-CM & \texttt{Gokul-waterlabs/ICD-10-CM} & Medical classifications & 74K \\
\bottomrule
\end{tabular}}
\caption{Dataset specifications for numerical embedding analysis across natural language contexts.}
\label{tab:lan-context-datasets}
\end{table*}

We provide an overview of the datasets we use for natural language context experiments in \Cref{tab:lan-context-datasets}.
In \Cref{sec:multitok}, we use BBC data as processed by \citet{Li_Guerin_Lin_2024}, which we retrieved from \href{https://huggingface.co/datasets/RealTimeData/bbc_news_alltime}{\texttt{RealTimeData/bbc\_news\_alltime}}.

In all cases, we trust the original creators of the datasets for handling private and sensitive information according to standard. As we do not redistribute the data, we take no specific steps towards verifying this point.

\section{Use of LLMs in writing and coding}
LLM-based technology was used to polish writing and jump-start coding. Authors take full responsibility for the contents of this paper. 

\end{document}